\documentclass[runningheads]{llncs}
\usepackage{graphicx}

\usepackage[table,dvipsnames]{xcolor}
\usepackage{tikz}
\usepackage{comment}
\usepackage{amsmath,amssymb} 
\usepackage{color}
\usepackage{wrapfig}
\usepackage{multirow}
\usepackage{floatrow}
\usepackage[normalem]{ulem}

\usepackage[accsupp]{axessibility}  

\usepackage{booktabs}
\usepackage{etoolbox}
\usepackage[format=plain,labelformat=simple,labelsep=period,font=small,compatibility=false]{caption}
\usepackage[font=footnotesize,skip=3pt,subrefformat=parens]{subcaption}

\newtoggle{comments}
\togglefalse{comments}
\iftoggle{comments}{%
    \newcommand{\tarasha}[1]{{\leavevmode\color{magenta}[Tarasha: #1]}}
    \newcommand{\peiyun}[1]{{\leavevmode\color{red}[Peiyun: #1]}}
    \newcommand{\achal}[1]{{\leavevmode\color{orange}[Achal: #1]}}
    \newcommand{\david}[1]{{\leavevmode\color{purple}[David: #1]}}
    \newcommand{\deva}[1]{{\leavevmode\color{blue}[Deva: #1]}}
    \newcommand{\newtext}[1]{{\leavevmode\color{red}#1}}
}{%
  \newcommand{\tarasha}[1]{}
  \newcommand{\peiyun}[1]{}
  \newcommand{\achal}[1]{}
  \newcommand{\david}[1]{}
  \newcommand{\deva}[1]{}
  \newcommand{\newtext}[1]{{\leavevmode\color{black}#1}}
}

\newfloatcommand{capbtabbox}{table}[][\FBwidth]
\usepackage[pagebackref,breaklinks,colorlinks]{hyperref}

\usepackage[capitalize]{cleveref}
\crefname{section}{Sec.}{Secs.}
\Crefname{section}{Section}{Sections}
\Crefname{table}{Table}{Tables}
\crefname{table}{Tab.}{Tabs.}

\def\etal{\textit{et al.}}
\definecolor{ForestGreen}{RGB}{34,139,34}

\begin{document}
\pagestyle{headings}
\mainmatter
\def\ECCVSubNumber{1105}  

\title{Differentiable Raycasting for Self-supervised Occupancy Forecasting} 

\titlerunning{Differentiable Raycasting for Self-supervised Occupancy Forecasting}
%
\author{Tarasha Khurana\inst{1}\thanks{equal contribution} \and
Peiyun Hu\inst{2}$^{*}$ \and
Achal Dave\inst{3} \and
Jason Ziglar\inst{2} \and
David Held\inst{1,2} \and
Deva Ramanan\inst{1,2}}
\authorrunning{T. Khurana$^*$, P. Hu$^*$, A. Dave, J. Ziglar, D. Held, D. Ramanan}
%
\institute{Carnegie Mellon University \and Argo AI \and Amazon }
\maketitle

\begin{abstract}

Motion planning for safe autonomous driving requires learning how the environment around an ego-vehicle evolves with time. Ego-centric perception of driveable regions in a scene not only changes with the motion of actors in the environment, but also with the movement of the ego-vehicle itself. Self-supervised representations proposed for large-scale planning, such as ego-centric freespace, confound these two motions, making the representation difficult to use for downstream motion planners. In this paper, we use \textit{geometric occupancy} as a natural alternative to view-dependent representations such as freespace. Occupancy maps naturally disentagle the motion of the environment from the motion of the ego-vehicle. However, one cannot directly observe the full 3D occupancy of a scene (due to occlusion), making it difficult to use as a signal for learning. Our key insight is to use \textit{differentiable raycasting} to ``render" future occupancy predictions into future LiDAR sweep predictions, which can be compared with ground-truth sweeps for self-supervised learning. The use of differentiable raycasting allows occupancy to {\em emerge} as an internal representation within the forecasting network. In the absence of groundtruth occupancy, we quantitatively evaluate the forecasting of raycasted LiDAR sweeps and show improvements of upto 15 F1 points. For downstream motion planners, where emergent occupancy can be directly used to guide non-driveable regions, this representation relatively reduces the number of collisions with objects by up to 17\% as compared to freespace-centric motion planners.

\end{abstract}

\section{Introduction}
\label{sec:intro}

To navigate in complex and dynamic environments such as urban cores, autonomous vehicles need to perceive actors and predict their future movements. Such knowledge is often represented in some form of forecasted occupancy~\cite{sadat2020perceive}, which downstream motion planners rely on to produce safe trajectories. When tackling the tasks of perception and prediction, standard solutions consist of perceptual modules such as object detection, tracking, and trajectory forecasting, which require a massive amount of object track labels. Such solutions do not scale given the speed that log data is being collected by large fleets.

\begin{figure}[t!]
    \centering
  \includegraphics[width=0.65\linewidth]{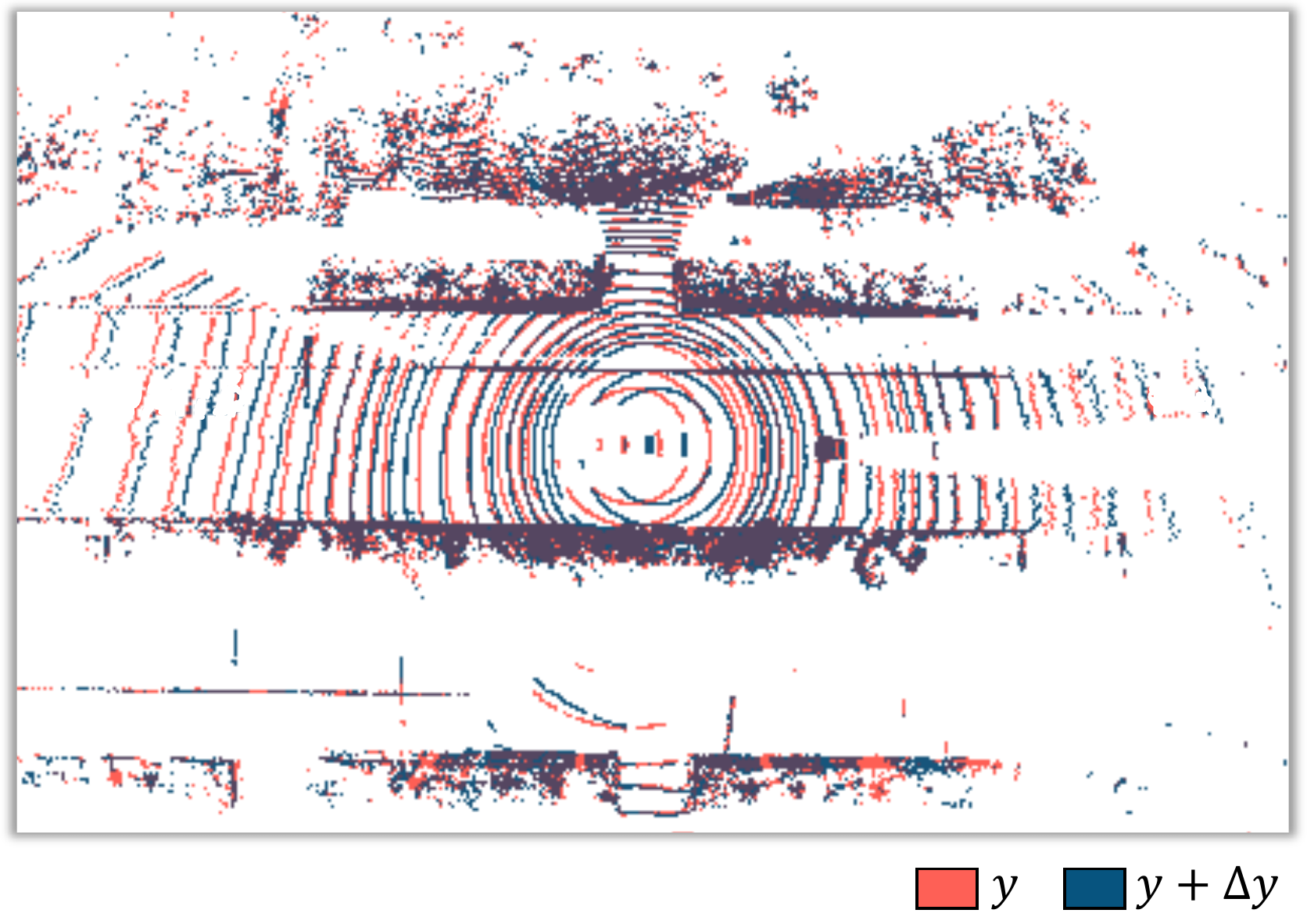}
  \caption{We pose-align two succesive LiDAR sweeps of a static scene $s$ to a common world coordinate-frame (using the notation of Fig.~\ref{fig:splash}). Even though there is zero scene motion $\Delta s$, points appear to drift or {\em swim} across surfaces. This is due to the fact that points are obtained by intersecting rays from a moving sensor $\Delta y$ with static scene geometry. This in turn implies that points can appear to move since they are not tied to physical locations on a surface. This apparent movement (${\Delta \tilde s}$)
  is in general a complex nonlinear transformation, even when the sensor motion $\Delta y$ is a simple translation (as shown above).  
  Traditional methods for self-supervised LiDAR forecasting \cite{weng2020inverting,wilson2021argoverse,Mittal_2020_CVPR,hu2021safe} require predicting the complex transformation ${\Delta \tilde s}$ which depends on the unknown $\Delta y$, while our differentiable-raycasting framework assumes $\Delta y$ is an {\em input}, dramatically simplifying the task of the forecasting network. From a planning perspective, we argue that the future (planned) change-in-pose {\em should} be an input rather than an output. 
  }
  \label{fig:swim}
\end{figure}

{\bf Freespace versus occupancy:} To avoid the need for costly human annotations, and to enable learning at scale, self-supervised representations such as ego-centric freespace \cite{hu2021safe} have been proposed.
However, such a representation couples the motion of the world with the motion of the ego-vehicle (Fig. \ref{fig:splash}).
Our key innovation in this paper is to learn an ego-pose independent and explainable representation for safe motion planning, which we call \textit{emergent occupancy}. Emergent occupancy decouples ego motion and scene motion using differentiable raycasting: we design a network that learns to ``space-time complete'' the future volumetric state of the world (in a world-coordinate frame) given past LiDAR observations. 
Consider an ego-vehicle that moves in a static scene. Here, LiDAR returns (even when aligned to a world-coordinate frame) will still {\em swim} along the surfaces of the fixed scene (Fig. \ref{fig:swim}). This implies that even when the world is static, most of what the ego-vehicle observes through the LiDAR sensor appears to move with complex nonlinear motion, but in fact those observations can be fully explained by static geometry and ego-motion (via raycasting). LiDAR forecasters need to implicitly predict this ego-motion of the car to produce accurate future returns. However, we argue that such prediction doesn't make sense for autonomous agents that {\em plan} their future motion. Importantly, our differentiable raycasting network has access to future camera ego-poses as {\em input}, both during training (since they are available in archival logs) and testing (since state-of-the-art planners explicitly search over candidate trajectories).


{\bf Self-supervision:} Note that ground-truth future volumetric occupancy is largely unavailable without human supervision, because the full 3D world is rarely observed; the ego-vehicle only sees a limited number of future views as recorded in a single archival log. To this end, we apply a differentiable raycaster that projects the forecasted volumetric occupancy into a LiDAR sweep, as seen by the future ego-vehicle motion in the log. We then use the difference between the raycasted sweep and actual sweep as a signal for self-supervised learning, allowing us to train models on massive amounts of unannotated logs. 

{\bf Planning:} Lastly, we show that such forecasted space-time occupancy can be jointly learned with space-time costmaps for end-to-end motion planning. Owing to LiDAR self-supervision, we are able to train on recent unsupervised LiDAR datasets~\cite{mao2021one} that are orders of magnitude larger than their annotated counterparts, resulting in significant improvement in accuracy for both forecasted occupancy and motion plans. Interestingly, as we increase the amount of archival training data at the cost of zero additional human annotation, object shape, tracks, and multiple futures ``emerge'' in the arbitrary quantities predicted by our model despite there being no direct supervision on ground-truth occupancy.

\section{Related Work}



%
%
%
%


{\bf Occupancy as a scene representation:} Knowledge regarding what is around an autonomous vehicle (AV) and what will happen next is captured in different representations throughout the standard modular perception and prediction (P\&P) pipeline~\cite{lang2019pointpillars,weng2019baseline,chai2019multipath,sadat2019jointly}. 
Instead of separate optimization of these modules~\cite{urmson2008autonomous,montemerlo2008junior}, Sadat et al.~\cite{sadat2020perceive} propose bird's-eye view (BEV) {\it semantic occupancy} that is end-to-end optimizable.
As an alternative to {\it semantic occupancy}, Hu et al.~\cite{hu2020you} propose BEV {\it ego-centric freespace} that can be self-supervised by raycasting on aligned LiDAR sweeps. However, the ego-centric freespace entangles motion from other actors, which is arguably more relevant for motion planning, with ego-motion.
In this paper, we propose {\it emergent occupancy} to isolate motion of other actors. While we focus on self-supervised learning at scale, we acknowledge that for motion planning, some semantic labelling is required (e.g., state of a traffic light) which can be incorporated via semi-supervised learning.



{\bf Differentiable raycasting:} Differentiable raycasting has shown great promise in learning the underlying scene structure given samples of observations for downstream novel view synthesis~\cite{mildenhall2020nerf}, pose estimation~\cite{yen2020inerf}, etc. In contrast, our application is best described as ``space-time scene completion'', where we learn a network to predict an explicit space-time occupancy volume. Furthermore, our approach differs from existing approaches in the following ways. We use LiDAR sequences as input and raycast LiDAR sweeps given future occupancy and sensor pose. We work with explicit volumetric representations ~\cite{lombardi2019neural} for dynamic scenes with a feed-forward network instead of test-time optimization~\cite{park2021nerfies}.

{\bf Self-supervision:} Standard P\&P solutions do not scale given how fast log data is collected by large fleets and how slow it is to curate object track labels. To enable learning on massive amount of unlabeled logs,
supervision from simulation~\cite{dosovitskiy2017carla,chen2020learning,codevilla2018end,codevilla2019exploring}, auto labeling using multi-view constraints~\cite{qi2021offboard}, and self-supervision have been proposed. Notably, tasks that can be naturally self-supervised by LiDAR sweeps e.g., scene flow~\cite{Mittal_2020_CVPR} have the potential to generalize better as they can leverage more data. More recently, LiDAR self-supervision has been explored in the context of point cloud forecasting~\cite{weng2020100k,weng2020inverting,wilson2021argoverse}. However, when predicting future sweeps given the history, as stated before, past approaches often tend to couple motion of the world with the motion of the ego-vehicle~\cite{weng2020100k}.

{\bf Motion Planning:} An understanding of what is around an AV and what will happen next~\cite{urmson2008autonomous} is crucial. This is typically done in the bird's eye-view (BEV) space by building a modular P\&P pipeline.
Although BEV motion planning does not precisely reflect planning in the 3D world, it is widely used as the highest-resolution and computation- and memory-efficient representation~\cite{zeng2019end,sadat2020perceive,casas2021mp3}. However, training such modules often requires a massive amount of data.
End-to-end learned planners requiring less human annotation have emerged,
with end-to-end imitation learning (IL) methods showing particular promise~\cite{codevilla2018end,rhinehart2018deep,chen2020learning}. Such methods often learn a neural network to map sensor data to either action (known as behavior cloning) or ``action-ready'' cost function (known as inverse optimal control)~\cite{osa2018algorithmic}.
However, they are often criticized for lack of explainable intermediate representations, making them less accountable for safety-critical applications~\cite{pomerleau1989alvinn}. More recently, end-to-end learned but modular methods producing explainable representations, e.g., neural motion planners~\cite{zeng2019end,sadat2020perceive,casas2021mp3} have been proposed. However, these still require costly object track labels. Unlike them, our approach
learns explainable intermediate representations that are explainable quantities for safety-critical motion planning without the need of track labels.

\section{Method}



Autonomous fleets provide an abundance of {\it aligned} sequences of LiDAR sweeps $\bf x$ and ego vehicle trajectories $\bf y$. How can we make use of such data to improve perception, prediction, and planning? In the sections to follow, we first define occupancy. Then we describe a self-supervised approach to predicting future occupancy. Finally, we describe an approach for integrating this forecasted occupancy into neural motion planners. Note that in the text that follows, we use ego-centric freespace and freespace interchangeably.

\subsection{Occupancy}


We define occupancy as the state of occupied space at a particular time instance. We use $\bf z$ to denote the true occupancy, which may not be directly observable due to visibility constraints. 
Let us write
\begin{equation}
    {\bf z[u]} \in \{0, 1\}, {\bf u} = (x, y, t), {\bf u \in U}
\end{equation}
to denote the occupancy of a voxel $\bf u$ in the space-time voxel grid $\bf U$, which can be {\it occupied} (1) or {\it free} (0). The spatial index of $\bf u$, i.e., $(x, y)$ represents the spatial location from a bird's-eye view.
Given a sequence of {\it aligned} sensor data and ego-vehicle trajectory $\bf (x, y)$, there may be multiple plausible occupancy states $\bf z$ that ``explain'' the sensor measurements. We denote this set of plausible occupancy states as $\bf Z$.

\textbf{Forecasting Occupancy.}  Suppose we split an aligned sequence of LiDAR sweeps and ego-vehicle trajectory $(\bf x, y)$ into a historic pair $(\bf x_1, y_1)$ and a future pair $(\bf x_2, y_2)$. Our goal is to learn a function $f$ that takes historical observations $(\bf x_1, y_1)$ as input and predicts emergent future occupancy $\bf \hat{z}_2$. Formally,
\begin{align}
    {\bf \hat{z}_2} = f(\bf x_1, y_1), \label{eq:occupancy_forecast}
\end{align}
If the true occupancy $\bf z_2$ were observable, we could directly supervise our forecaster, $f$.
Unfortunately, in practice, we only observe LiDAR sweeps, $\bf x$. We show in the next section how to supervise $f$ with LiDAR sweeps using differentiable raycasting techniques.

\subsection{Raycasting}
Given an occupancy estimate $\bf \hat{z}$, sensor origin $\bf y$ and directional unit vectors for rays $\bf r$, a differentiable raycaster $\mathcal{R}$ can raycast LiDAR sweeps $\bf \hat x$.
We use $\bf \hat d$ to represent the expected distance these rays travel before hitting obstacles: $ {\bf \hat d} = \mathcal{R} \bf(r; \hat z, y)$.
Then we can reconstruct the raycast LiDAR sweep $\bf \hat x$ as $\bf \hat x =  y + \hat d * r$.

\subsection{Learning to Forecast Occupancy}

\begin{figure}[t]
\centering
\includegraphics[width=0.65\linewidth]{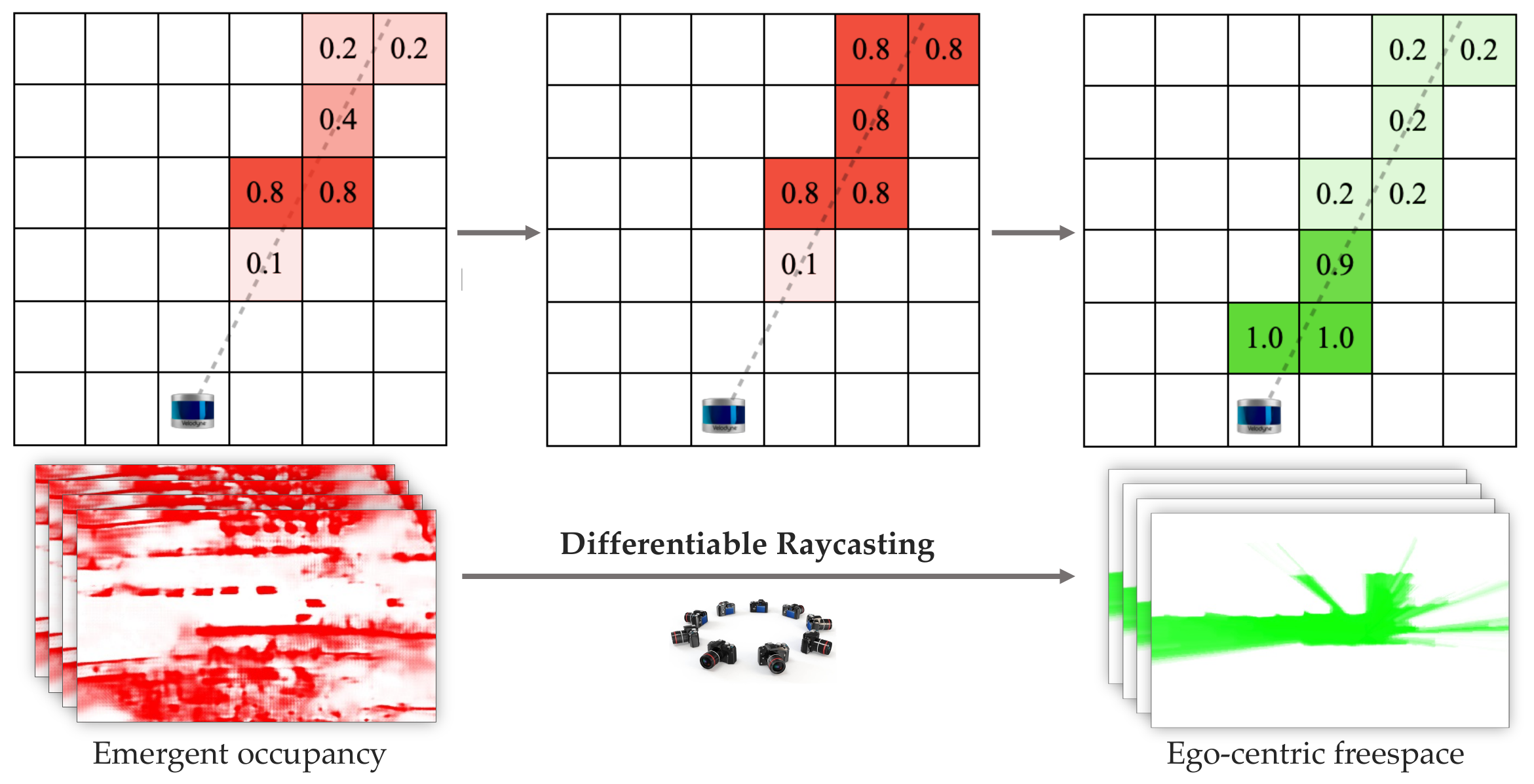}
  \caption{Differentiable procedure for estimating ego-centric freespace from volumetric occupancy, necessary for computing the loss from \eqref{eq:loss}. The left image depicts predicted emergent occupancy, on which we perform a cumulative max along the LiDAR ray from known sensor poses (middle), which is differentiable because it is essentially re-indexing. The result is then inverted to produce (soft) visible ego-centric freespace estimates. To identify BEV pixels along the LiDAR ray, we perform fast voxel traversal in 2D~\cite{amanatides1987fast}.
  }
  \label{fig:diff-render}
\end{figure}




Given the predicted occupancy $\bf \hat{z}_2$ (Eq.~\ref{eq:occupancy_forecast}), and the captured sensor pose $\bf y_2$, a differentiable raycaster $\mathcal{R}$ can take rays $\bf r_2$ as input and produce
${\bf \hat{d}_2} = \mathcal{R}(\bf r_2; \hat{z}_2, y_2)$. Note that this formulation allows us to decouple the motion of the world captured by change in occupancy, $\bf \hat{z}_2$, and the motion of the ego-vehicle captured by change in sensor origin, $\bf y_2$.

This also allows us to supervise $\bf \hat{z}_2$ using a loss function that measures the difference between the raycast distance $\bf \hat{d}_2$ and the ground-truth distance $\bf d_2$.
\begin{equation}
    L_r = \text{loss} \bf (\hat d_2,d_2) \label{eq:loss}
\end{equation}



{\bf Loss function:} One natural loss function might be distance between the raycast depth and measured depth along each ray. In practice, we care most about disagreements of freespace which can inform safe motion plans. To emphasize such disagreements, we define voxels encountered along the ray as having a free versus not-free binary label, and use a binary cross-entropy loss (summed over all voxels encountered by each ray until the boundary of voxel grid, ref. \newtext{Fig.~\ref{fig:diff-render}}). We adopt an encoder-decoder architecture that predicts future emergent occupancy given historical LiDAR sweeps, differentiably raycasts future LiDAR sweeps and self-supervises using archival sweeps \newtext{(ref. highlighted branch of Fig. \ref{fig:plan_arch} (a))}.





\subsection{Learning to Plan}
The previous section described an approach for predicting future LiDAR returns via differentiable raycasting of BEV space-time occupancy maps. We now show that such costmaps can be integrated directly into an end-to-end motion planner that makes use of space-time costmaps for scoring candidate trajectories. We follow~\cite{hu2021safe}, but modify their derivation to take into account emergent occupancy.

{\bf Max-margin planning:} We learn a model $g$ to predict a space-time cost map, $\bf c_2$, over future timestamps given past observations $\bf (x_1, y_1)$:
\begin{equation}
    \label{eq:costmap}
    {\bf c_2} = g(\bf x_1, y_1),\text{\ where\ }\bf c_2[u] \in \mathbb{R}, u \in U_2
\end{equation}
where $\bf U_2$ represents the space-time voxel grid over future timestamps. We define the cost of a trajectory as the sum of costs at its space-time way-points. The best candidate future trajectory according to the cost map is the one with the lowest cost:
\begin{equation}
    {\bf \hat y_2^*} = \arg \min_{\bf \hat y \in Y_2} C(\bf \hat y; c_2) = \arg \min_{\bf \hat y \in Y_2} \sum_{u \in \hat y} c_2[u]
\end{equation}
where $\bf Y_2$ represents the set of viable future trajectories.

{\bf Loss function:} We use a max-margin loss function, where the target cost of a candidate trajectory ($\bf \hat y$) is equal to the cost of the expert trajectory ($\bf y_2$) plus a margin. We can write the objective as follows:
\begin{equation}
    L_p = \left[C\left({\bf y_2; c_2}\right) - \left(\min_{\bf \hat y \in Y_2} C\left({\bf \hat y; c_2}\right) - D\left({\bf \hat y, y_2}\right) \right) \right]_+
\end{equation}
where $[\cdot]_+=\max(\cdot, 0)$ and $D$ is a function that quantifies the desired margin between the cost of a candidate trajectory and the cost of an expert trajectory. A common choice for $D$ is Euclidean distance between pairs of way-points:
\begin{equation}
    D{\bf (\hat y_2, y_2) = ||\hat y_2, y_2||}_2 \label{eq:margin}
\end{equation}

\let\oldFBaskip\FBaskip
\let\oldFBbskip\FBbskip
\renewcommand\FBaskip{-2.3em}
\renewcommand\FBbskip{-2.0em}
\begin{wrapfigure}{r}{0.15\linewidth}
  \begin{center}
  \includegraphics[width=\linewidth]{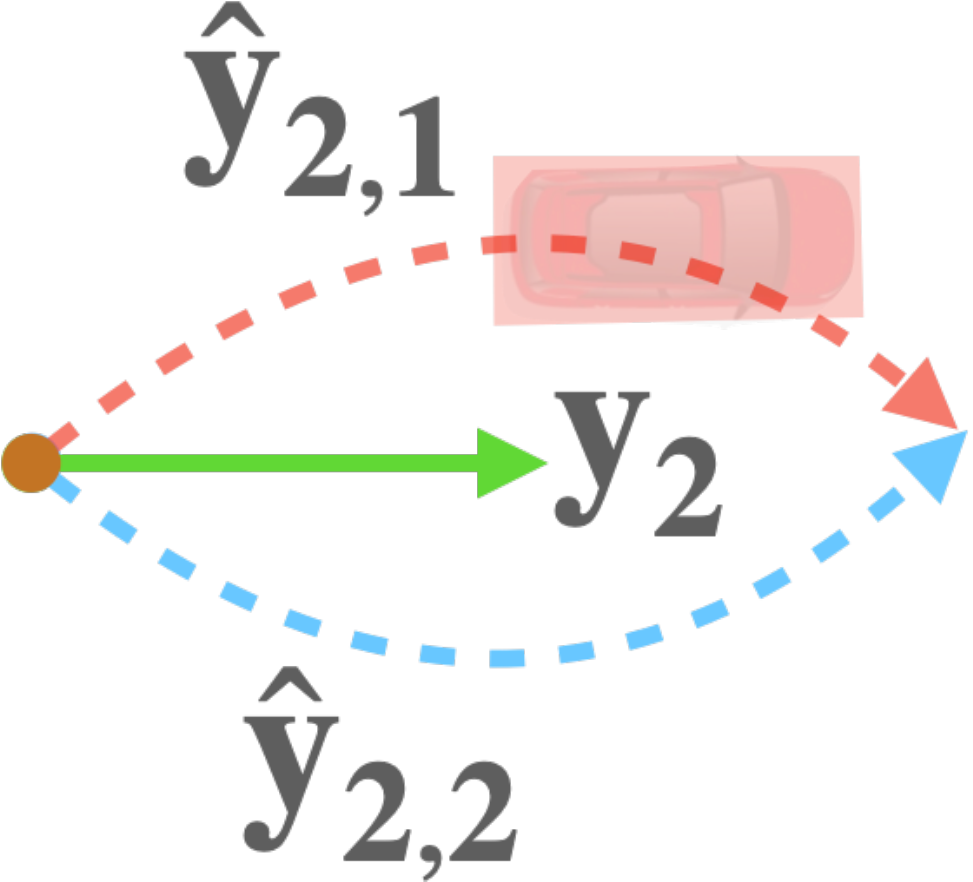}
  \label{fig:incentive}
  \end{center}
\end{wrapfigure}
\renewcommand\FBaskip{\oldFBaskip}
\renewcommand\FBbskip{\oldFBbskip}
Learning cost maps that reflect such cost margins only requires expert demonstrations, which are readily available in archival log data. However, sometimes candidates trajectories that are equally distant from the expert one should bear different costs. We provide an example (right) where the red trajectory should cost more than blue in the presence of an obstacle despite both being equidistant from the expert demonstration.

{\bf Guided planning:} To further distinguish among candidate trajectories, one could introduce extra penalty terms given additional supervision.
\begin{equation}
    D{\bf (\hat y_2, y_2) = ||\hat y_2, y_2||}_2 + \gamma~P(\bf \hat y_2) \label{eq:guided-margin}
\end{equation}
where $P$ represents a penalty function and $\gamma$ is a predefined scaling factor. Zeng et al.~\cite{zeng2019end} propose to define an additional penalty such that candidate trajectories that collide with object boxes would cost an additional $\gamma$ in addition to the deviation from the expert demonstration. We refer to this approach as {\it object-guided planning}, which is effective but costly as it requires object track labels.




More scalable alternatives to object supervision can be adopted, such as formulation of the penalty term proposed by Hu et al.~\cite{hu2021safe}. Concretely, candidate trajectories that reach outside the freespace as observed by future LiDAR poses would incur an additional penalty. We refer to this as {\it freespace-guided planning}. 

\begin{figure*}[t!]
    \centering
    \begin{subfigure}[t]{\textwidth}
        \centering
        \includegraphics[width=\linewidth]{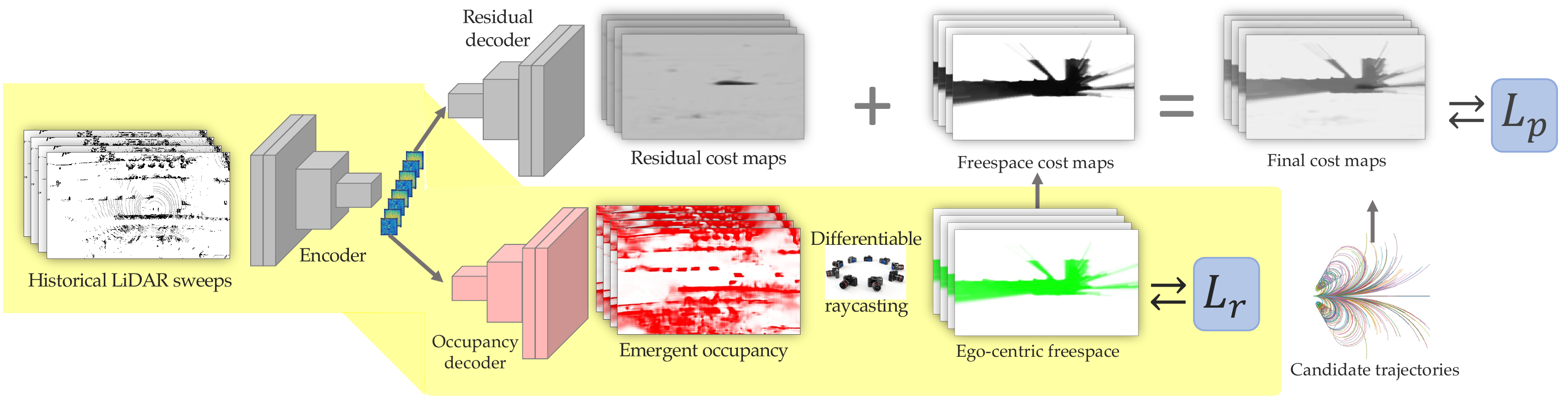}
        \caption{Training architecture.}
    \end{subfigure}%
    \\
    \begin{subfigure}[t]{0.83\textwidth}
        \centering
        \includegraphics[width=\linewidth]{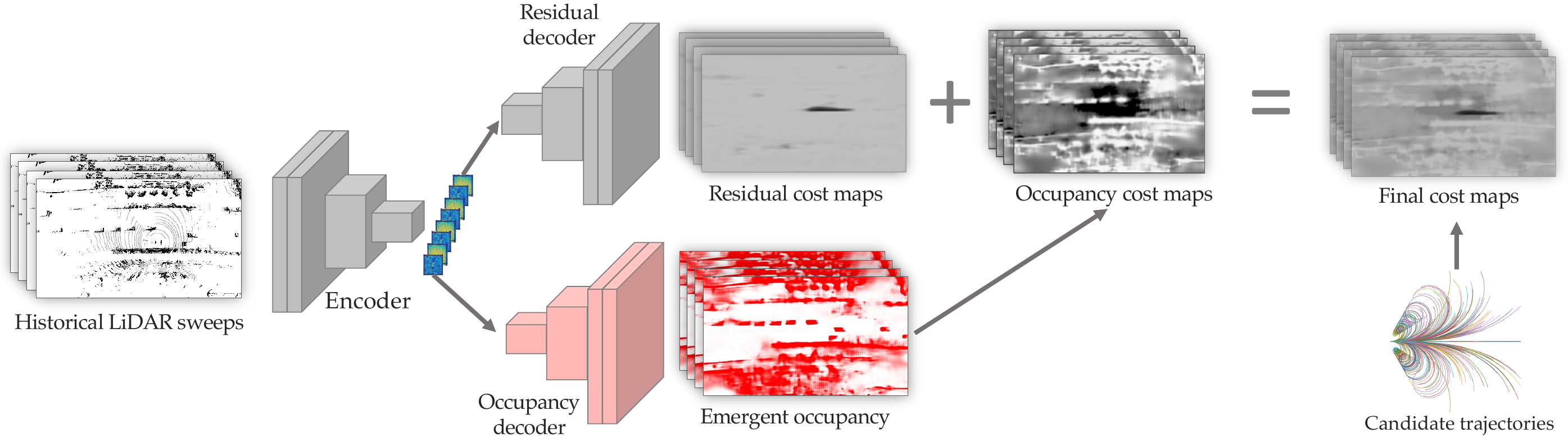}
        \caption{Inference architecture.}
    \end{subfigure}
    \caption{\newtext{Overview of our training and inference-time planning architectures. Highlighted network branch in (a) is used to learn future emergent occupancy, which is augmented by the residual branch that predicts residual cost maps, eventually used in computing a guided planning loss.
    }
    \tarasha{does anyone think the inference arch should also have the candidate trajectories icon?} \deva{I'd vote for it!}
    }
    \label{fig:plan_arch}
\end{figure*}

{\bf Residual costmaps:} Instead of directly predicting the cost map $\bf c_2[u]$, we follow prior work \cite{hu2021safe} and predict a residual cost map $\bf \Tilde{c}_2[u]$ that is added to the cost map from freespace estimate based on predicted emergent occupancy.
\begin{equation}
    \label{eq:residual}
    \bf c_2[u] = \Tilde{c}_2[u] + \alpha~\text{proj}(\hat z_2; y_2)[u],~ u \in \hat{y}_2
\end{equation}
where $\alpha$ is a predefined constant and $\bf \Tilde{c}_2$ represents the predicted residual cost map. The operation $\text{proj}(\bf \hat z_2; y_2)$ is illustrated in Fig.~\ref{fig:diff-render}.

{\bf Multi-task planning ({\color{red}new}):} In addition to the raycasting loss in Fig. \ref{fig:plan_arch}, we add $L_p$ as an additional planning loss. 
In other words, the emergent occupancy prediction architecture is augmented with another decoder branch to predict the residual cost maps while sharing the encoder features. Because of this, emergent occupancy forecasting becomes the auxiliary task for the end-to-end motion planner. We illustrate the network architecture during training in Fig.~\ref{fig:plan_arch} (a). 

{\bf Test-time occupancy cost maps ({\color{red}new}):} At test time, to compute ego-centric freespace cost maps based on predicted emergent occupancy, for each candidate sample trajectory, one would need to perform raycasting from its waypoints, which is prohibitively expensive. Fortunately, this is exactly equivalent to directly accessing emergent occupancy on the waypoints along the candidate trajectory (because of the cumulative max-operation used in deriving freespace from occupancy - see Fig.~\ref{fig:diff-render}), as formally expressed in Eq.~\eqref{eq:equivalence}. 
\begin{equation}
    \text{proj}\bf(\hat{z}_2; \hat{y}_2)[u] = \hat{z}_2[u],~u \in \hat{y}_2 \label{eq:equivalence}
\end{equation}



The simplified test-time architecture is illustrated in Fig.~\ref{fig:plan_arch} (b). When optimizing for future trajectories, we restrict the search space of future trajectories to the ones with a smooth transition from the past trajectory~\cite{zeng2019end,hu2021safe}.  
Please refer to the supplement for other implementation information such as detailed network architecture.

\section{Experiments}




\noindent {\bf Datasets:} We evaluate occupancy forecasting and motion planning on two datasets: nuScenes~\cite{caesar2019nuscenes} and ONCE~\cite{mao2021one}. nuScenes features real-world driving data with 1,000 fully annotated 15 second logs.
ONCE is the largest driving dataset with 150 hours of real-world data including 1 million LiDAR sweeps, collected in a range of diverse environments such as urban and suburban areas.
As annotation is expensive, only a small subset of logs in ONCE are fully annotated, making it ideal for self-supervised learning. 
We include comparison against state-of-the-art forecasting and planning approaches on both datasets. We also construct multiple baselines for all ablative evaluation for bird's eye-view motion planning.
To understand how our occupancy forecasting and motion planning performance scales to an increasing amount of training data, we randomly curate different training sets of the datasets. Since only a small subset of 8K samples in ONCE is labeled, we do this by progressively increasing the number of training samples by adding scenes from both their labeled and unlabeled-small splits, which include 8K, and 86K training samples respectively. Some of our analysis exists only on the combined labeled and unlabeled-small split which totals to 94K samples. For nuScenes, we randomly sample scenes from their official training set. For all experiments that follow, we take in a historical LiDAR stack of 2 seconds and forecast for the next 3 seconds.

\subsection{Emergent Occupancy Forecasting}
\label{sec:freespace-forecast-eval}


\begin{table}[t]
  \centering
 \resizebox{0.75\linewidth}{!}{
  \begin{tabular}{@{}cccccc@{}}
  \toprule
  	Dataset & Diff. Raycast & $~\frac{|\mathbf{d} - \hat{\mathbf{d}}|}{\mathbf{d}} (\downarrow)$ & $~$ BCE ($\downarrow$) & $~$ F1 ($\uparrow$) & $~$ AP ($\uparrow$)\\
	\midrule
	\multirow{2}{*}{nuScenes} & - \cite{hu2021safe} & 0.297 & 0.221 & 0.665 & 0.769 \\
	& $\checkmark$ & \textbf{0.242} & \textbf{0.140} & \textbf{0.777} & \textbf{0.863}\\
	\midrule
	\multirow{2}{*}{ONCE} & - \cite{hu2021safe} & 0.371 & 0.143 & 0.635 & 0.732\\
	& $\checkmark$ &\textbf{0.243}& \textbf{0.097} & \textbf{0.787} & \textbf{0.827} \\
    \bottomrule
  \end{tabular}
  }
  \caption{Indirect evaluation of emergent occupancy forecasting  with respect to groundtruth LiDAR sweeps. On both nuScenes and ONCE, we significantly improve forecasting accuracy across all metrics by using differentiable raycasting for decoupling the scene and ego-motion, unlike Hu \textit{et al.} \cite{hu2021safe}.
  \label{tab:occ}
  }
\end{table}

\begin{figure*}[t!]
    \centering
    \begin{subfigure}[t]{0.32\textwidth}
        \centering
        \includegraphics[width=\linewidth]{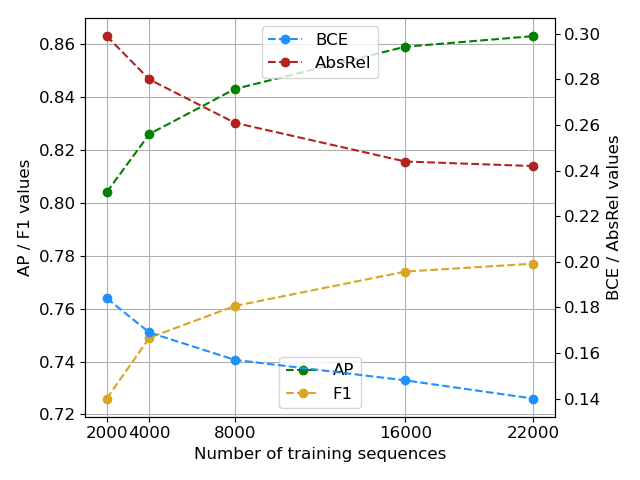}
        \caption{nuScenes}
    \end{subfigure}%
    ~
    \begin{subfigure}[t]{0.32\textwidth}
        \centering
        \includegraphics[width=\linewidth]{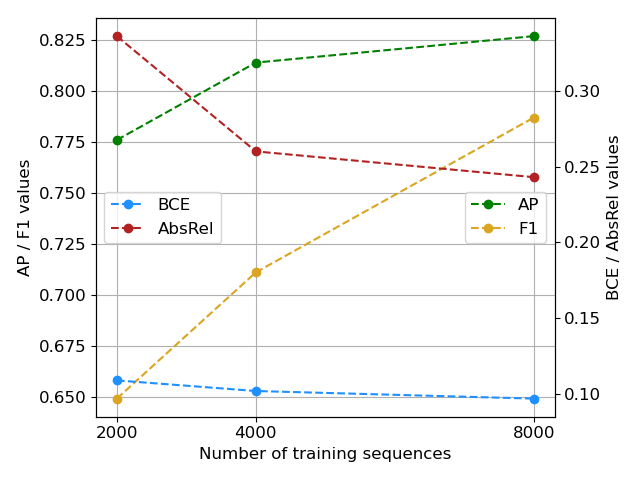}
        \caption{ONCE-labeled}
    \end{subfigure}
    ~
    \begin{subfigure}[t]{0.32\textwidth}
        \centering
        \includegraphics[width=\linewidth]{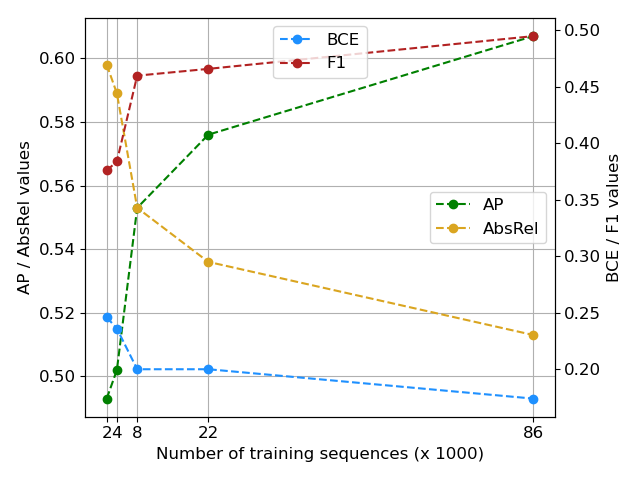}
        \caption{ONCE-unlabeled}
    \end{subfigure}%
    \caption{We highlight the merits of our self-supervised approach which can be given any amount of unlabeled LiDAR data to train on, in the form of posed archival LiDAR sweeps, thereby increasing the performance of emergent occupancy forecasting (evaluated using classification metrics such as average-precision and F1). Please refer to the supplement for corresponding tables.
    }
    \label{fig:graphs_occ}
\end{figure*}

\noindent {\bf Metrics:} Since, the groundtruth for true occupancy is unavailable, we quantitatively evaluate the LiDAR sweeps raycast from the emergent occupancy predictions. Specifically, our first evaluation computes the absolute relative error between the groundtruth distance traveled by every ray starting from the sensor origin, and the expected distance traveled by corresponding rays; where the expected distance is obtained by casting rays through the forecasted occupancy. Second, we score \textit{every} BEV voxel traversed by a ray using its `free' or `not-free' state. This dense per-ray evaluation is equivalent to evaluating the per-pixel binary classification of an ego-centric freespace map with respect to its groundtruth, allowing us to compare to the baseline discussed below. We compute the dense binary cross-entropy, average precision and the F1-score. All metrics are averaged across all prediction timesteps (up to 3s).

\noindent {\bf Baseline:} We re-implement the future-freespace architecture from \cite{hu2021safe} which directly forecasts ego-centric freespace. For building our architecture, we adapt this network to predict an arbitrary quantity which differentiably raycasts into ego-centric freespace given a sensor location. On training this architecture in a self-supervised manner, the arbitrary quantity \textit{emerges} into emergent occupancy, an explainable intermediate representation for downstream motion planners.

\noindent {\bf Main results:} We compare the performance of both approaches in Tab. \ref{tab:occ}. Note the drastic improvement in all metrics on using differentiable raycasting to decouple the scene motion from the ego-motion of the sensor on both nuScenes and ONCE. With increase of up to 15\% F1 points, we highlight the high-quality of our predicted occupancy and the pronounced effect of adding differentiable raycasting.
Our results show that occupancy reasoning is an important intermediate task, \textit{even} if the end-goal is simply understanding freespace: Our method, which predicts occupancy as an intermediate target, outperforms \cite{hu2021safe}, which directly aims to predict freespace. \newtext{Fig. \ref{fig:qual} visualizes predicted ego-centric freespace for a single scenario in ONCE using \cite{hu2021safe} and our approach at $t = {0,3}s$ in the future. In Fig. \ref{fig:graphs_occ}, we show how adding more training samples to both datasets result in an upward trend in performance across \textit{all} metrics. This increasing generalizability and scaling of training data comes for free with our self-supervised approach.

\begin{figure}
    \centering
    \includegraphics[width=0.7\linewidth]{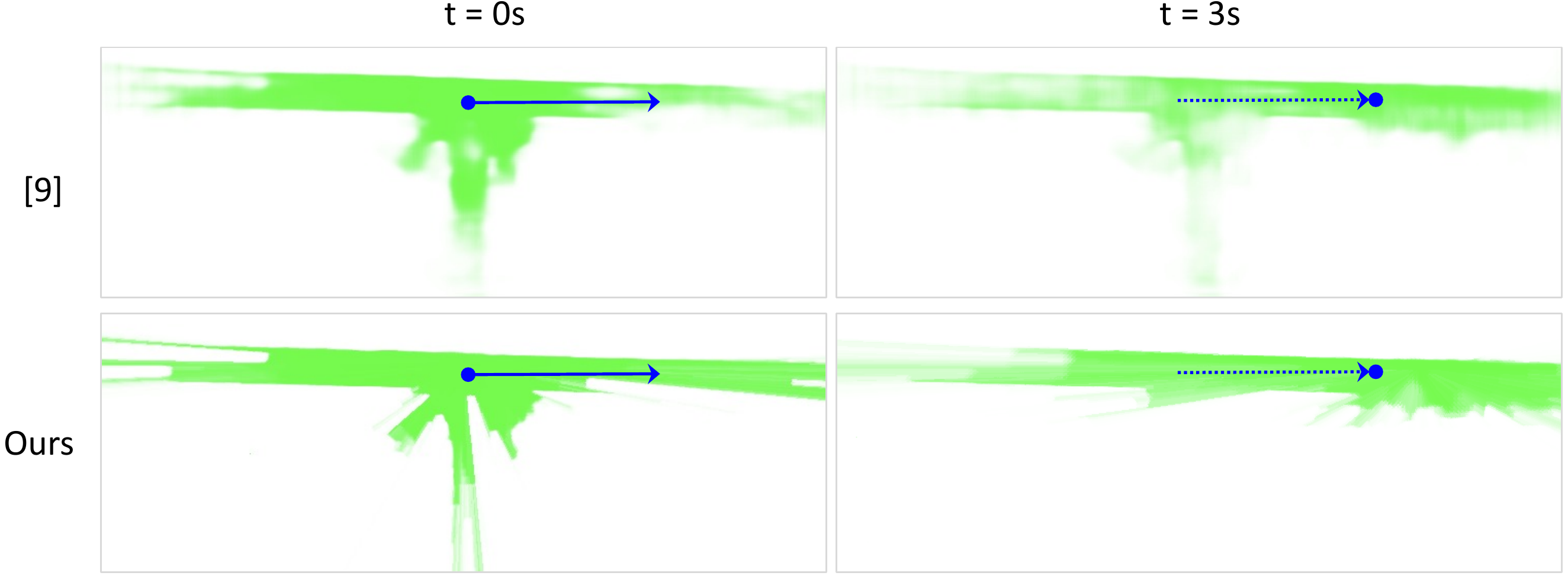}
    \caption{\newtext{Future ego-centric freespace from~\cite{hu2021safe} and our model, raycasted from predicted emergent occupancy. Note how the presence of moving and parked cars on roadsides is captured well by our approach even 3s in the future.
    }}
    \label{fig:qual}
\end{figure}}


\subsection{Motion Planning}

{\bf Metrics:} We follow prior works and compute three metrics for evaluating motion planning performance, including (1) L2 error; (2) point collision rate; (3) box collision rate. The L2 distance measures how close the planned trajectory follows the expert trajectory at each future timestamp. The point collision rate measures how often the planned waypoint is within the BEV boxes of other objects. The box collision rate measures how often the BEV box of the ego-vehicle intersects with BEV boxes of other objects.

\let\oldFBaskip\FBaskip
\let\oldFBbskip\FBbskip
\renewcommand\FBaskip{-2.3em}
\renewcommand\FBbskip{-2.0em}
\begin{wrapfigure}[9]{r}{0.25\textwidth}
    \begin{center}
    \includegraphics[width=\linewidth]{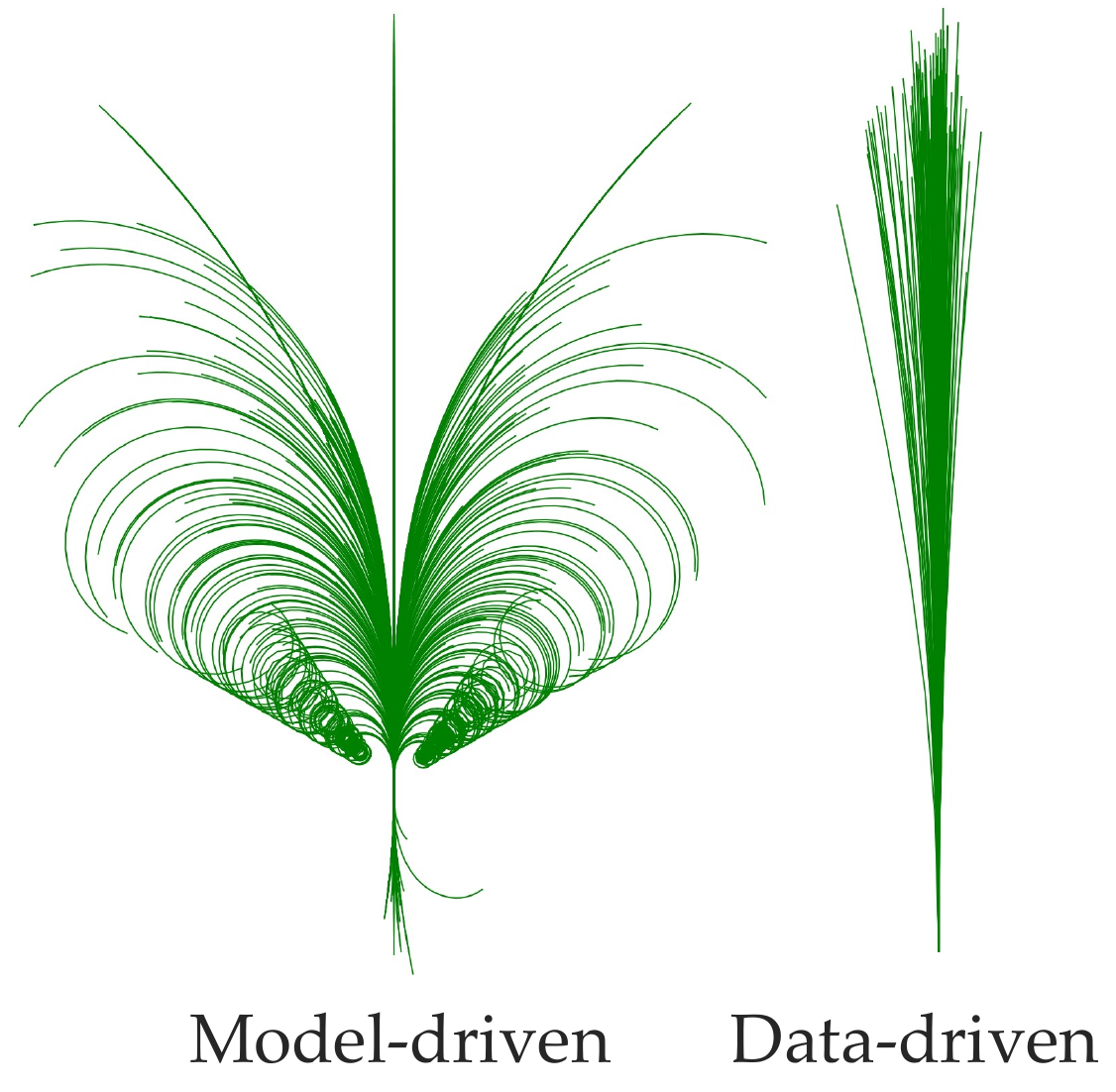}
    \label{fig:trajectories}

    \end{center}
\end{wrapfigure}
\renewcommand\FBaskip{\oldFBaskip}
\renewcommand\FBbskip{\oldFBbskip}
\noindent {\bf Trajectory sampling:} When evaluating performance on nuScenes, we follow
previous state-of-the-art approaches~\cite{zeng2019end,hu2021safe} and sample a combination of straight lines, circles, and clothoid curves as trajectory samples. Owing to the scene diversity in ONCE, we notice that such a sampling strategy does not capture the distribution of expert trajectories on ONCE as they range widely in their velocities and directions. Inspired by~\cite{casas2021mp3}, we sample a data-driven trajectories to complement the model-driven samples (right). The supplement provides more details on our data-driven sampler.

\subsection*{Planning on nuScenes}

\begin{figure}[t]
  \begin{floatrow}
  \capbtabbox{%
  \raisebox{-.5\height}{
  \resizebox{0.97\linewidth}{!}{
      \begin{tabular}{@{}lcccccc@{}}\toprule
    	\multirow{2}{*}{nuScenes} & \multicolumn{3}{c}{Box Collision (\%)} & \multicolumn{3}{c}{L2 Error (m)} \\
        \cmidrule(lr){2-4} \cmidrule(lr){5-7}
        & $1s$ & $2s$ & $3s$ & $1s$ & $2s$ & $3s$\\
        \midrule
        IL~\cite{ratliff2006maximum} & 0.08 & 0.27 & 1.95 & \textbf{0.44} & \textbf{1.15} & \textbf{2.47} \\
        FF~\cite{hu2021safe} & 0.06 & 0.17 & 1.07 & 0.55 & 1.20 & 2.54 \\
        {\bf \color{red} Ours} & \textbf{0.04} & \textbf{0.09} & \textbf{0.88} & 0.67 & 1.36 & 2.78\\
        \midrule
        NMP~\cite{zeng2019end} & 0.04 & 0.12 & \textit{\underline{0.87}} & \textit{\underline{0.53}} & \textit{\underline{1.25}} & \textit{\underline{2.67}} \\
        P3~\cite{sadat2020perceive} & \textit{\underline{0.00}} & \textit{\underline{0.05}} & 1.03 & 0.59 & 1.34 & 2.82 \\
        \bottomrule
      \end{tabular}
  }
  }
  }{%
      \caption{We compare end-to-end state-of-the-art motion planners on nuScenes-val. 
      NMP and P3 are supervised approaches
      that have access to object tracking labels.
    \label{tab:nuscenes}
      }
  }
  \ffigbox{%
  \centering
    \includegraphics[width=0.85\linewidth]{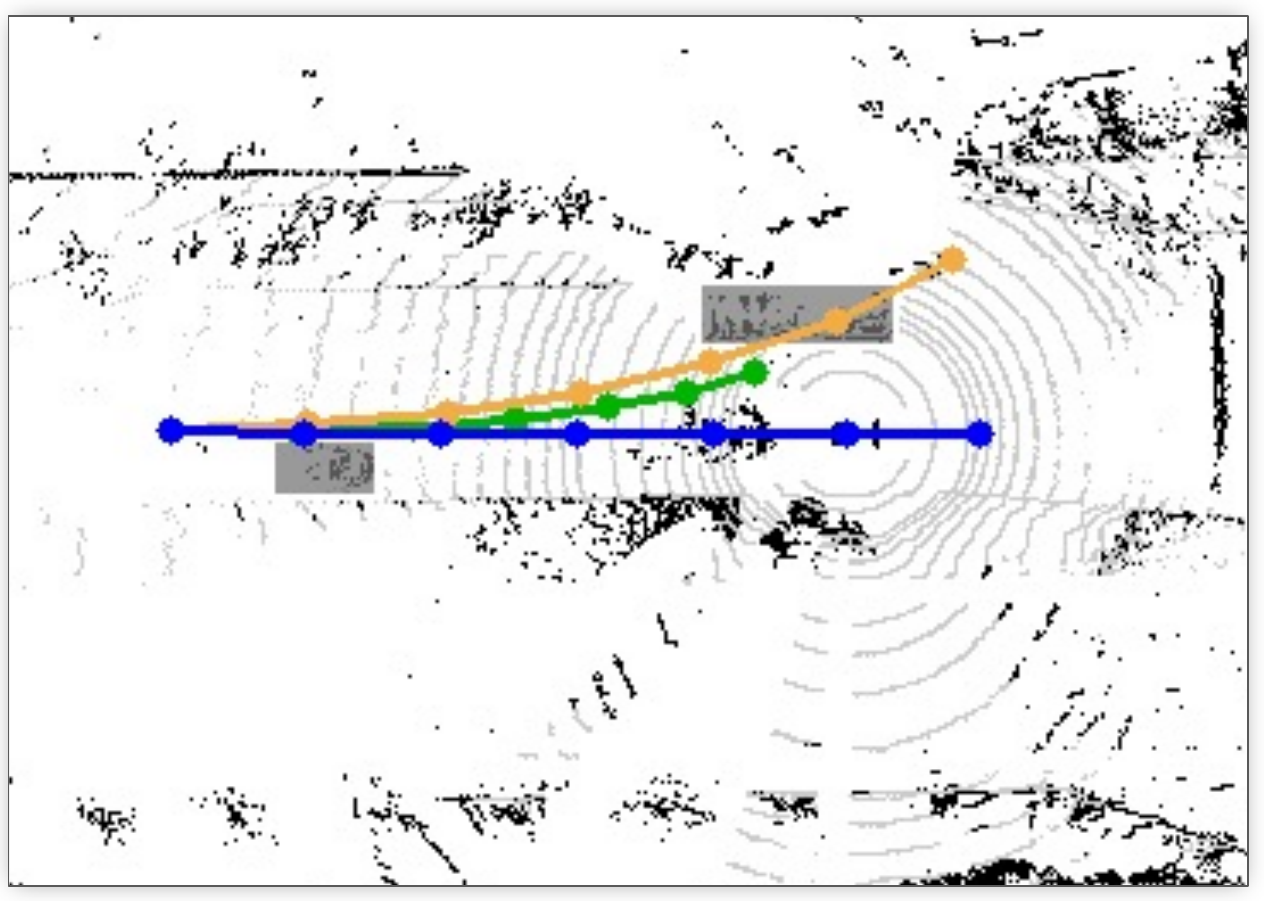}
  }{%
    \caption{ A \textcolor{Dandelion}{vanilla spacetime trajectory} with a lower L2 error wrt. \textcolor{Blue}{expert}, may collide into \textcolor{gray}{objects} unlike a \textcolor{Green}{proposed trajectory} with larger L2 error but no collision. }
    \label{fig:l2error}

  }
\end{floatrow}
\end{figure}

\begin{table*}
  \centering
  \begin{tabular}{@{}ccccccccccccc@{}}\toprule
	\multirow{2}{*}{} & Freespace & Multi & Diff. & \multicolumn{3}{c}{Box Collision (\%)} & \multicolumn{3}{c}{Point Collision (\%)} & \multicolumn{3}{c}{L2 Error (m)} \\
    \cmidrule(lr){5-7} \cmidrule(lr){8-10} \cmidrule(lr){11-13}
    & Guided & Task & Raycast & $~~1s~~$ & $~~2s~~$ & $3s$ & $~~~1s~~$ & $~~2s~~~$ & $3s$ & $~~~1s~~$ & $~~2s~~~$ & $3s$\\
	\midrule
	(a) & - & - & - & 0.08 & 0.27 & 1.95 & \textbf{0.00} & \textbf{0.00} & 0.35 & 0.44 & 1.15 & 2.47 \\
	(b) & $\checkmark$ & - & - & 0.06 & 0.17 & 1.07 & \textbf{0.00} & 0.01 & 0.04 & 0.55 & 1.20 & 2.54 \\
	(c) & - & $\checkmark$ & - & 0.08 & 0.17 & 1.29 & \textbf{0.00} & 0.02 & 0.08 & \textbf{0.42} & \textbf{1.06} & \textbf{2.30} \\
	(d) & $\checkmark$ & $\checkmark$ & - & \textbf{0.02} & 0.10 & 1.10 & \textbf{0.00} & \textbf{0.00} & 0.08 & 0.52 & 1.22 & 2.64  \\
	{\bf \color{red}(e)} & $\checkmark$ & $\checkmark$ & $\checkmark$ & 0.04 & \textbf{0.09} & \textbf{0.88} & \textbf{0.00} & 0.01 & \textbf{0.03} & 0.67 & 1.36 & 2.78 \\
    \bottomrule
  \end{tabular}
  \caption{Ablation studies on nuScenes-val. Note that (a) is IL, (b) is FF, and (e) is {\color{red} \bf Ours} in Tab.~\ref{tab:nuscenes}.
  \label{tab:nuscenes_ablation}
  }
\end{table*}
{\bf Baselines:} We compare our proposed approach to four baseline end-to-end motion planners. First, we implement a pure imitation learning (IL) baseline, a max-margin neural motion planner self-supervised by expert trajectories, as described in Eq.~\eqref{eq:margin}. Second, we re-implement future-freespace-guided max-margin planner (FF) proposed by Hu \textit{et al.}~\cite{hu2021safe}, as captured by Eq.~\eqref{eq:guided-margin}. Third, we re-implement a simplified neural motion planner (NMP) without modeling costs related to map information and traffic light status as such information is unavailable on nuScenes. Last, we re-implement a simplified version of perceive, predict, and plan (P3) where we do not distinguish semantic occupancy of different classes. To ensure a fair comparison, we adopt the same neural net architecture for the baselines and our approach.

\noindent {\bf Main results:} As Tab.~\ref{tab:nuscenes} shows, in terms of collision rates, our self-supervised approach outperforms both self-supervised baselines (IL and FF) by a large margin. Moreover, our approach achieves the same collision rate at 3s as the best of supervised baselines. We also observe a commonly observed trade-off between L2 errors and collision rates~\cite{zeng2019end}. For example, pure imitation learning achieves the lowest L2 errors with the highest collision rates.

\noindent {\bf Ablation studies:} We perform extensive ablation studies \newtext{in Tab.~\ref{tab:nuscenes_ablation}} to understand where improvements come from.
There are three main observations:
\begin{itemize}
    \item Differentiable raycasting reduces collision rate at further horizon (3s), as seen in (d) vs. (e), suggesting decoupling motion of the world (space-time occupancy) from ego-motion is helpful when learning long range cost maps.
    \item Multi-task learning further reduces collision rates, as seen in (a) vs. (c). Training max-margin planners with an auxiliary self-supervised forecasting task significantly reduces the collision rates without hurting L2.
    \item Freespace-guided cost margin is crucial to lowering collision rates, as seen in (a) vs. (b), (c) vs. (d). However, there is a trade-off: the L2 errors tend to increase as being expert-like (at all costs) is no longer the only objective. In Fig. \ref{fig:l2error}, we show an example result describing why L2 error is a misleading metric that  doesn't allow for alternate future plans that are otherwise viable. \newtext{Additionally, Casas \emph{et al.}~\cite{casas2021mp3} show that collision rate is a more consistent metric between evaluation in the open- and closed-loop setups.}
\end{itemize}
\begin{figure*}[t]
    \centering
    \includegraphics[width=\linewidth]{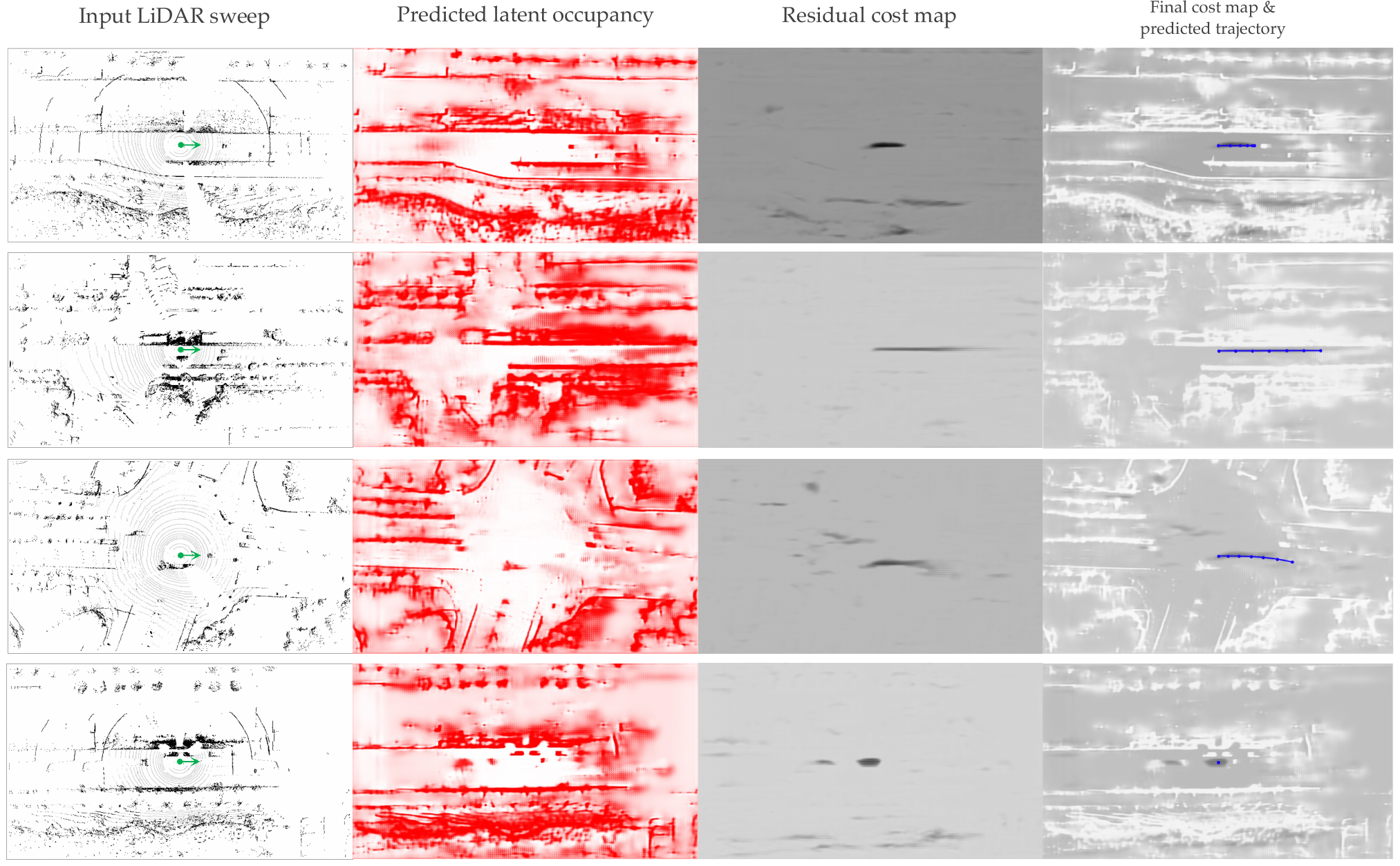}
    \caption{Qualitative results of our learned model. From top to bottom, we visualize various scenarios, including slowing down, speeding up, \newtext{navigating an intersection} and staying still.
    All columns after the first one are visualized at future timestamp t=0.5s. We successfully forecast the motion of surrounding objects, e.g. in third row, which results in safer planned trajectories.
    }
    \label{fig:qualitative}
\end{figure*}
\begin{figure}[t]
    \centering
  \includegraphics[width=0.7\linewidth]{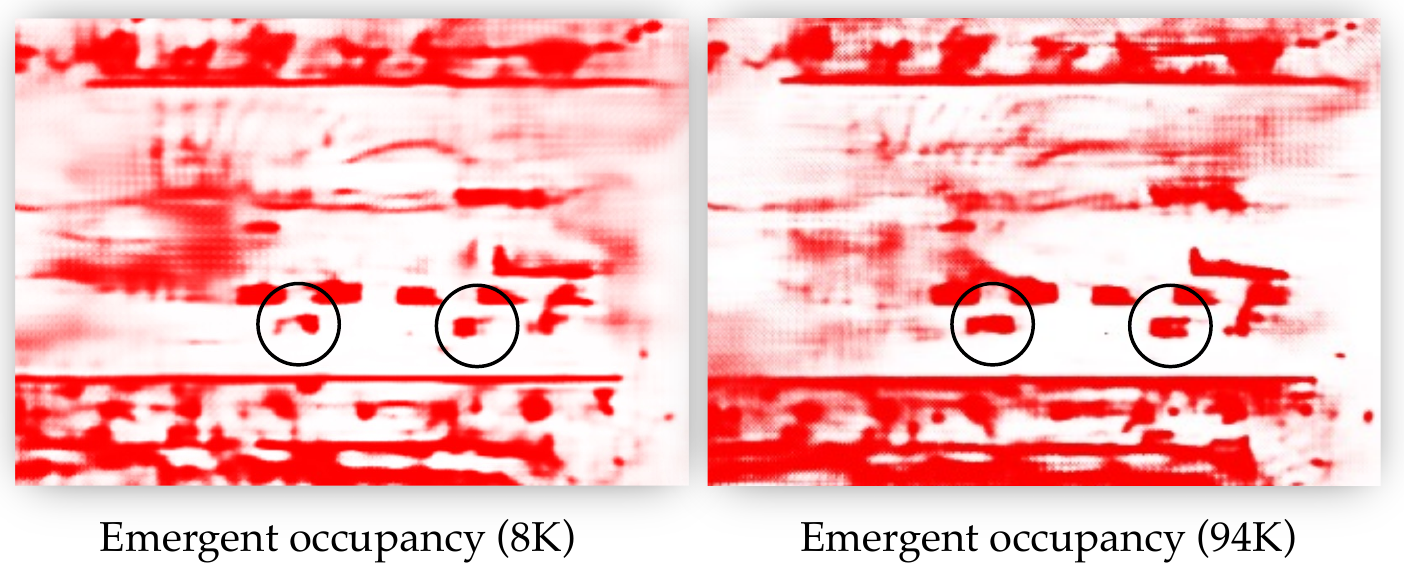}
  \caption{Evolution of estimated emergent future occupancy.
  } 
  \label{fig:objectshape-more-data-helps}
\end{figure}
%
\subsection*{Planning on ONCE}

{\bf Baseline:} ONCE offers a massive amount of unlabeled, diverse LiDAR sweeps paired with ego-vehicle trajectories and a small fully labeled subset of about 8K samples. We train a re-implemented neural motion planner as a supervised baseline on the fully labeled subset. We train our self-supervised approach over a wide range of training sizes, from 2K to 94K.


\let\oldFBaskip\FBaskip
\let\oldFBbskip\FBbskip
\renewcommand\FBaskip{-3.5em}
\renewcommand\FBbskip{0.0em}
\begin{wrapfigure}[14]{r}{0.45\textwidth}
    \begin{center}
    \includegraphics[width=\linewidth]{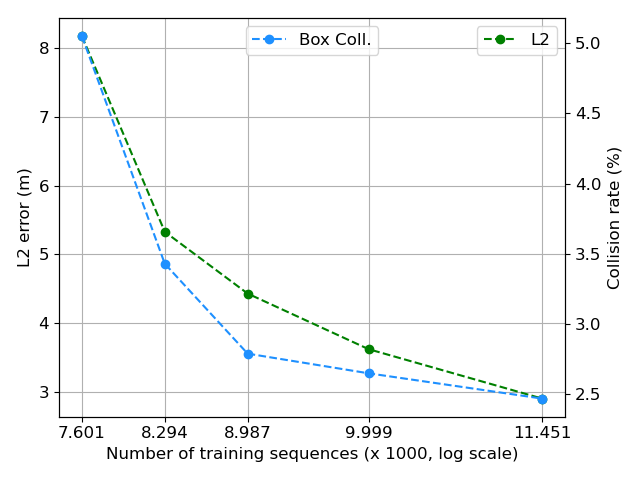}
    \caption{Planning performance vs. larger ONCE training set size. 
    }
    \label{fig:scale}
    \end{center}
\end{wrapfigure}
\renewcommand\FBaskip{\oldFBaskip}
\renewcommand\FBbskip{\oldFBbskip}

\noindent {\bf Main results:} Perhaps unsurprisingly, our first observation is that the metrics on ONCE are inflated as compared to nuScenes, because of the diverse range of environments ONCE features, ranging from straight highways to complex city road structures. To show the scalability of our approach on such a diverse and large dataset, we plot the L2 error and (box) collision rate at 3s as a function of the amount of training data in Fig.~\ref{fig:scale}. Both the L2 error and the collision rate of our approach continue to improve as we increase the size of the training set. In comparison, the supervised neural motion planner achieves an L2 error of 4.45m and a box collision rate of 2.54\% at a training size of 8K.

At 94K training samples, our self-supervised approach achieves a dramatically lower L2 error of 2.9m and a lower collision rate of 2.47\%. Importantly, such scalability for motion planning comes for free as our approach is self-supervised. We show some qualitative results on the ONCE dataset in Fig. \ref{fig:qualitative} where our approach is able to deal with a number of varying driving scenarios; decelerate and stop when necessary, predict long trajectories when unoccupied regions are predicted ahead, avoid collisions with other vehicles while navigating an intersection, or stay stationary. Please refer to our supplement for further quantitative evaluation, visualization of future cost maps and more qualitative examples that feature failure cases (e.g., forecasted occupancy diffuses over time).

\noindent {\bf Evolution of occupancy estimates:} Our model tends to produce better estimates of emergent occupancy as we increase the amount of training data. The percent of semantic object pixels recalled from the ground-truth semantic object labels in our predicted occupancy map increases from 51\% to 59\% at t=0s when we increase the amount of training data from 8K to 94K. Qualitatively, this can be seen in Fig. \ref{fig:objectshape-more-data-helps} where the shape of two cars in the right lane looks more ``space-time complete'' for the model trained with increased data. 


\section{Conclusion}

We propose {\it emergent occupancy} as a self-supervised and explainable representation for motion planning. Our novel differentiable raycasting procedure enables the learning of occupancy forecasting under the self-supervised task of LiDAR sweep forecasting. The raycasting setup also allows us to decouple ego motion from scene motion, making forecasting an easier task for the network to learn.  Experimental results suggest that such decoupling is also helpful for downstream motion planning. Such training at scale allows object shape, tracks, and multiple futures to ``emerge'' in the predicted emergent occupancy. \\ 



\noindent {\bf Acknowledgments} This work was supported by the CMU Argo AI Center for Autonomous Vehicle Research.

\clearpage

\begin{center}
    {\large \textbf{Appendix}}
\end{center}

\appendix
In this supplement, we discuss more details of our experimental setup in Sec. \ref{sec:expsetup}, discuss supplementary evaluation of occupancy forecasting in Sec. \ref{sec:occ} and analyse the quantitative and qualitative performance of our motion planning architecture further in Sec. \ref{sec:once}.

\section{Experimental Setup}
\label{sec:expsetup}

\subsection{Network Architecture}

\paragraph{Architecture Implementation}
We use the same neural network architecture as proposed by Zeng \etal~\cite{zeng2019end} and developed on by Hu \etal~\cite{hu2021safe}. Different from these two networks, we use two decoders, one that predicts the emergent occupancy cost maps, and one that predicts the residual cost maps. The differentiable raycaster proposed by us acts as a layer over the occupancy cost maps, that produces raycast sweeps for 7 future timesteps (accounting for three seconds in the future).

Freespace is computed from these sweeps and it is used in 3 places in the network: (1) in computing a dense per-pixel classification loss with the groundtruth freespace, (2) in computing the final cost maps which are a sum of the freespace and the residual cost maps, and (3) in computation of the cost margin for the planning loss.

\paragraph{Input and output} We follow the same input and output BEV data format as that used in Hu \etal~\cite{hu2021safe} for nuScenes, except that we now take input from [-2s, 0s]. For nuScenes, this means 20 input timestamps and a stack of $704 \times 400 \times 20$ size. For ONCE, since the LiDAR sweeps are collected at 2Hz, this stack is of size $704 \times 400 \times 5$ to accomodate 2s of input data. The output of the network for both the datasets is of size $704 \times 400 \times 7$ to account for 3s of forecasts at a 0.5s interval, starting at the 0th timestep. \newtext{Each pixel in the BEV map covers an area of $0.2m \times 0.2m$.} To compute groundtruth freespace cost maps, we apply ground segmentation~\cite{hu2021safe,himmelsbach2010fast} to the output LiDAR sweep and raycast as described in the main paper.

\paragraph{Differentiable raycaster} First, we collect a set of rays, with origin as the position of the ego-vehicle in the world coordinate frame and endpoints as the endpoints in a groundtruth LiDAR sweep. For a given ray (origin and direction), we find the voxels that the ray travels in the BEV LiDAR scan using a fast voxel traversal algorithm proposed by Amanatides \etal \cite{amanatides1987fast}. Given all the voxels along a ray, we perform a soft raycast along the ray as follows: we sample occupancy states given the predicted occupancy probabilities, raycast to get free vs occluded space, and average the raycasts over all samples. In practice, we do this analytically by computing the expectation, as done by many prior works based on volume rendering ~\cite{mildenhall2020nerf}.

\subsection{Data-driven sampler}
Following prior work that uses data-driven trajectory sampling techniques for evaluating the performance in mapless driving scenarios~\cite{casas2021mp3}, we curate a dataset of expert trajectories by binning the trajectories in the train set by their velocity. Once this dataset is curated during preprocessing, we use retrieval based on the past timestep's speed and direction profiles to index into the appropriate bin in the dataset. When a velocity is not available in the set of data-driven trajectories, we compute the nearest speed and angle from the set, for a given sample. From this nearest bin, we randomly sample 200 valid trajectories and append them to our set of 2000 model-based trajectories.

\begin{figure}
    \centering
    \includegraphics[width=0.4\linewidth]{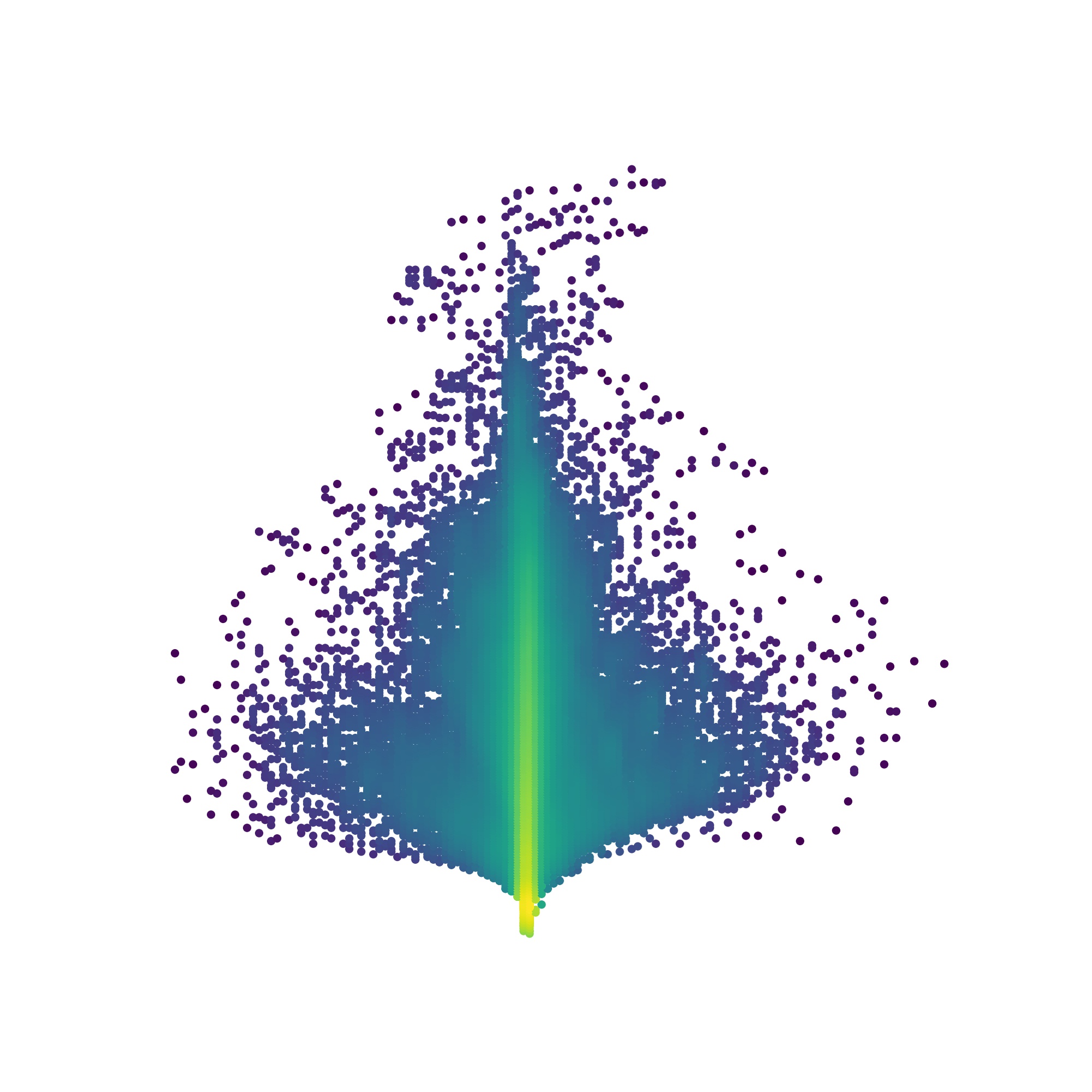}
    \includegraphics[width=0.4\linewidth]{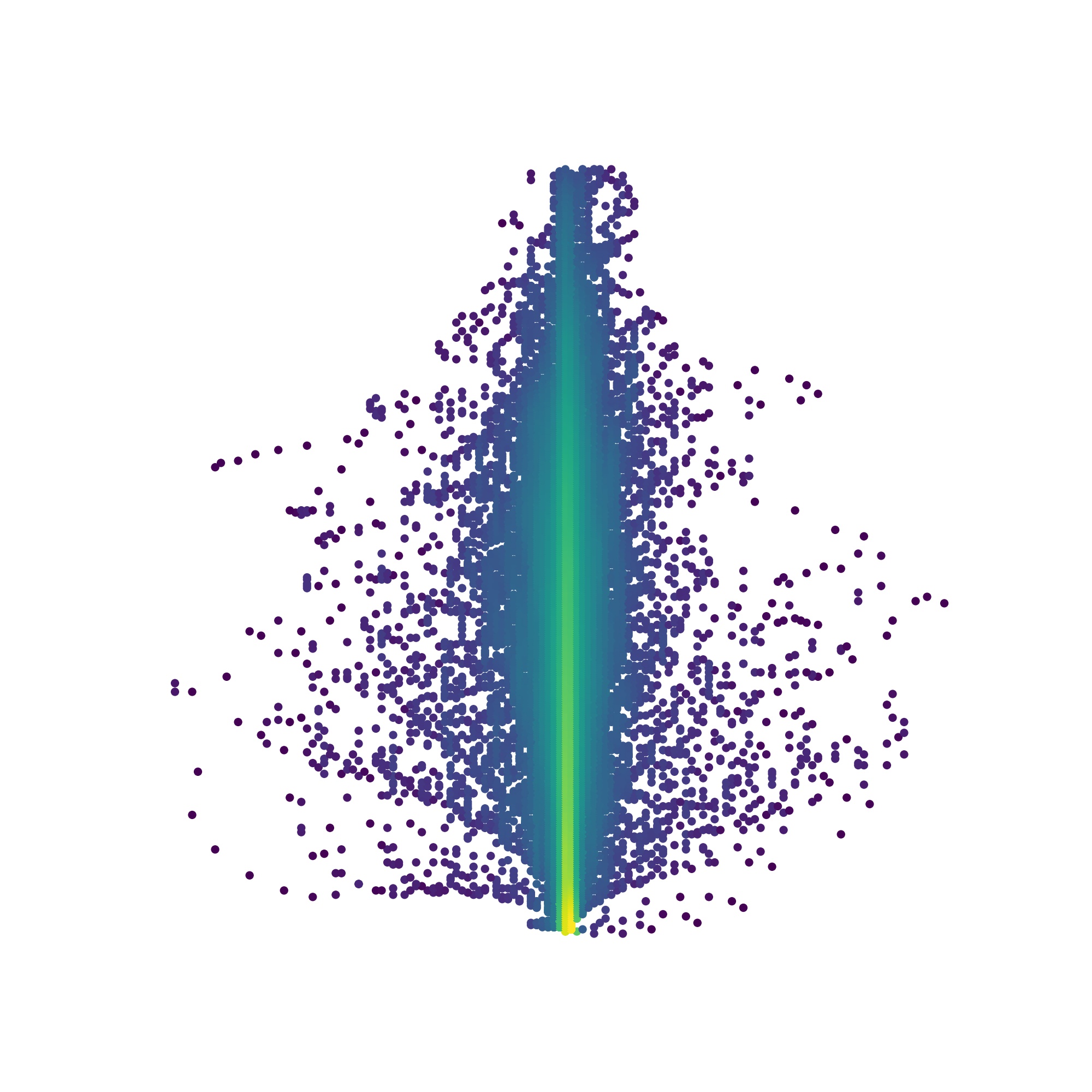}
    \caption{Distribution of train trajectories in nuScenes (\textbf{left}) and ONCE (\textbf{right}).}
    \label{fig:nuscvsonce}
\end{figure}

This approach avoids arbitrary choice of steering profiles for the ONCE dataset, since this information in unknown in ONCE (note that this is available for nuScenes with the CAN bus data). This is useful because in comparison to nuScenes, the ONCE dataset is composed of a complementary set of trajectories, as shown in Fig. \ref{fig:nuscvsonce}. Using this data-driven trajectory sampler in conjunction with the standard trajectory sampler gives us a complete coverage of possible future trajectories, including the ones that appear the most in the ONCE dataset.

\section{Occupancy Forecasting}
\label{sec:occ}

We supplement the evaluation of occupancy forecasting on ONCE and nuScenes in the main paper by providing complete results of Fig. 7 in Tab. \ref{tab:ffeval-nusc-once}. The unlabeled subsets of ONCE do not include samples from the labeled train set. As described, increasing the amount of training data directly impacts the improvement in performance. It is worth noting the performance difference between the 8k set of ONCE-labeled and ONCE-unlabeled. The higher metrics on the labeled set indicates that the ONCE labeled set is much higher quality and falls in the same data distribution as compared to the val set. nuScenes training subsets also show increasing performance with increase in data.

\begin{table}[t]
  \centering
 \resizebox{0.7\linewidth}{!}{
  \begin{tabular}{@{}cccccc@{}}
  \toprule
  	Dataset & Size & $~\frac{|\mathbf{d} - \hat{\mathbf{d}}|}{\mathbf{d}} (\downarrow)$ & $~$ BCE ($\downarrow$) & $~$ F1 ($\uparrow$) & $~$ AP ($\uparrow$)\\
	\midrule
	\multirow{3}{*}{ONCE} & 2,000 & 0.336	& 0.109 & 0.649 & 0.776\\
	 & 4,000 & 0.260	& 0.102 & 0.711 & 0.814\\
	 & 8,000 & \textbf{0.243} & \textbf{0.097} & \textbf{0.787} & \textbf{0.827}\\
	\midrule
	& 2,000 & 0.598 & 0.246 & 0.376 & 0.493\\
	ONCE& 4,000 & 0.589 & 0.236 & 0.384 & 0.502\\
	(unlabeled)& 8,000  & 0.553 & 0.200 & 0.460 & 0.553\\
	& 22,000 & 0.536 & 0.200 & 0.466 & 0.576\\
	& 86,000 & \textbf{0.513} & \textbf{0.174} & \textbf{0.495} & \textbf{0.607}\\
	\midrule
	 & 2,000 & 0.299	& 0.184	& 0.726	& 0.804\\
	 & 4,000 & 0.280	& 0.169	& 0.749	& 0.826\\
	nuScenes & 8,000 & 0.261	& 0.157	& 0.761	& 0.843\\
	 & 16,000 & 0.244 & 0.148 & 0.774 & 0.859\\
	 & 22,000 & \textbf{0.242} & \textbf{0.140} & \textbf{0.777} & \textbf{0.863}\\
    \bottomrule
  \end{tabular}
  }
  \vspace{3pt}
  \caption{Supplementary table for the evaluation of occupancy forecasting on ONCE-val and nuScenes-val with models trained on different subsets of the ONCE labeled, unlabeled and nuScenes train set.
  \label{tab:ffeval-nusc-once}
  }
\end{table}

\section{Motion Planning}
\label{sec:once}

\subsection{Planning on ONCE}

\paragraph{Quantitative Analysis}
This section supplements our results on the ONCE dataset from the main paper. We show the complete results of Fig. 12 in Tab. \ref{tab:once}. Note that as the amount of data is increased during training, the L2 error and box collision rate decreases dramatically. Even though box collision rate is a stricter metric than point collision rate, we see a consistent trend in it at the longest horizon. Our best model beats the neural motion planner~\cite{zeng2019end} baseline described in the main paper. Note that such a baseline can only be trained with the labeled training set of ONCE (with 8K samples), whereas all the raw unlabelled LiDAR logs in ONCE can be used by our method since it is self-supervised.

We also conduct an ablative study of our approach on the ONCE dataset in Tab. \ref{tab:ablations-once}. Note that since the hyperparameters are not tuned for the ONCE dataset, the best performing method on the Box Collision metric at 3s horizon is by Hu \textit{et al.} \cite{hu2021safe}. Intuitively, this difference in performance shows that the trajectories selected by our planner pass close to the objects in the environment, such that they incur a box collision but not point collision. This is expected as even though we outperform Hu \textit{et al.} \cite{hu2021safe} on occupancy forecasting, the guided planning loss used optimizes for point collision by summing per way-point occupancy cost instead of box collision, on which we outperform all other methods at 3s horizon.

\begin{table*}[t]
  \centering
  \begin{tabular}{@{}cccccccccc@{}}\toprule
	Training & \multicolumn{3}{c}{L2 Error (m)} & \multicolumn{3}{c}{Point Collision (\%)} & \multicolumn{3}{c}{Box Collision (\%)}\\
    \cmidrule(lr){2-4} \cmidrule(lr){5-7} \cmidrule(lr){8-10}
    size & $~~1s~~$ & $~~2s~~$ & $3s$ & $~~~1s~~$ & $~~2s~~~$ & $3s$ & $~~~1s~~$ & $~~2s~~~$ & $3s$\\
    \midrule
	8,000 & 0.84 & 2.26 & 4.45 & 0.00 & 0.04 & 1.06 & 0.04 & 0.14 & 2.54\\
	\midrule
	2,000 & 1.97 & 4.37 & 8.18 & 0.07 & 0.39 & 1.84 & 0.77 & 2.36 & 5.05 \\
	4,000 & 1.13 & 2.79 & 5.33 & 0.00 & 0.04 & 1.02 & 0.14 & 0.81 & 3.43 \\
	8,000 & 1.00 & 2.33 & 4.43	& 0.00 & 0.04 & \textbf{0.74} & 0.04 & \textbf{0.28} & 2.79\\
	22,000 & 0.71 & 1.87 & 3.62 & 0.00 & 0.14 & 1.06 & 0.04 & 0.39 & 2.65\\
    94,000 & \textbf{0.56} & \textbf{1.49} & \textbf{2.90} & \textbf{0.00} & \textbf{0.04} & 0.99 & \textbf{0.00} & 0.39 & \textbf{2.47}\\
    \bottomrule
  \end{tabular}
  \vspace{3pt}
  \caption{Planning metrics at different amount of training data on ONCE-val. First row corresponds to our reimplementation of the neural motion planner~\cite{zeng2019end} baseline described in the main paper.}
  \label{tab:once}
\end{table*}

\begin{table*}[t]
  \centering
  \resizebox{\linewidth}{!}{
  \begin{tabular}{@{}ccccccccccccc@{}}\toprule
	\multirow{2}{*}{} & Cost & Mid & Diff. & \multicolumn{3}{r}{L2 Distance (m)} & \multicolumn{3}{r}{Point Collision (\%)} & \multicolumn{3}{r}{Box Collision (\%)}\\
    \cmidrule(lr){5-7} \cmidrule(lr){8-10} \cmidrule(lr){11-13}
    & Margin & Task & Raycast & $~~1s~~$ & $~~2s~~$ & $3s$ & $~~~1s~~$ & $~~2s~~~$ & $3s$ & $~~~1s~~$ & $~~2s~~~$ & $3s$\\
    \midrule
	(a) & - & - & - & \textbf{0.61} & \textbf{1.64} & \textbf{3.33} & \textbf{0.00} & \textbf{0.00} & 1.02 & \textbf{0.00} & 0.42 & 2.47 \\
	 (b) & $\checkmark$ & - & - & 0.80 & 2.12 & 4.15 & \textbf{0.00} & \textbf{0.00} & 0.78 & \textbf{0.00} & \textbf{0.18} & \textbf{1.84} \\
	(c) & - & $\checkmark$ & - & 0.89 & 2.40 & 4.78 & \textbf{0.00} & 0.04 & 1.63 & \textbf{0.00} & 0.35 & 3.85 \\
	(d) & $\checkmark$ & $\checkmark$ & - & 0.90 & 2.49 & 4.99 & \textbf{0.00} & 0.04 & 1.10 & 0.07 & 0.25 & 2.61 \\
    {\bf \color{red}(e)} & $\checkmark$ & $\checkmark$& $\checkmark$ & 1.00 & 2.33 & 4.43 & \textbf{0.00} & 0.04 & \textbf{0.74} & 0.04 & 0.28 & 2.79\\
    \bottomrule
  \end{tabular}
  }
    \vspace{3pt}

  \caption{Ablation studies on ONCE. Note that (a) is IL, (b) is FF, and (e) is {\color{red} \bf Ours}.}
  \label{tab:ablations-once}
\end{table*}

\paragraph{Qualitative Analysis (5-min summary video)}
We show qualitative results of our approach in \href{https://youtu.be/QqSCu0KJ2FM}{this video}. The first example shows the input to our network, followed by a visualization of the predicted future {\color{red}\textbf{emergent/latent occupancy}}. Finally, we show the {\color{gray}\textbf{total cost maps}}, which are a sum of the occupancy and residual cost maps and the {\color{blue}\textbf{final output trajectory}}.

Additionally, we show three examples of the evolution of predicted emergent occupancy. In the first visual, note how object shapes are completed using prior knowledge collected during training. Second visual shows a straight moving car's tracks evolving with time. Finally, in the third example, for a car turning right at an intersection, notice how the multiple possible futures of the car grow into a triangular blob, indicating that according to the occupancy probability, the car could have moved at any angle, either moving straight across the intersection or turning right.

\subsection{Ablations on training architecture}

While we compute the predicted cost margin in the main paper by summing the egocentric-freespace cost maps with the residual cost maps for an apples-to-apples comparison with Hu \etal \cite{hu2021safe}, a more natural training architecture for motion planning would sum up the occupancy and residual cost maps during training, similar to the test-time architecture. Such an architecture would compute egocentric-freespace only for self-supervision with the multi-task loss and use the occupancy cost maps for motion planning. In Tab. \ref{tab:once}, we evaluate training with this ablated architecture. Note that since the occupancy cost maps are now optimized directly during training, the performance across all metrics increases.

\begin{table*}[t]
  \centering
  \begin{tabular}{@{}ccccccccccc@{}}\toprule
	\multirow{2}{*}{Dataset} & Training & \multicolumn{3}{c}{L2 Error (m)} & \multicolumn{3}{c}{Point Collision (\%)} & \multicolumn{3}{c}{Box Collision (\%)}\\
    \cmidrule(lr){3-5} \cmidrule(lr){6-8} \cmidrule(lr){9-11}
    & size & $~~1s~~$ & $~~2s~~$ & $3s$ & $~~~1s~~$ & $~~2s~~~$ & $3s$ & $~~~1s~~$ & $~~2s~~~$ & $3s$\\
    \midrule
	nuScenes & - & 0.76 & 1.61 & 3.23 & 0.00 & 0.00 & 0.15 & 0.04 & 0.15 & 0.98\\
	\midrule
     & 2,000 & 2.10 & 4.39 & 7.74 & \textbf{0.00} & 0.25 & 1.45 & 0.32 & 1.94 & 4.80\\
     & 4,000 & 1.09	& 2.73 & 5.15 & \textbf{0.00} & 0.04 & 0.99 & 0.07 & 0.53 & 3.28\\
	ONCE & 8,000 & 0.87 & 2.24 & 4.32 & \textbf{0.00} & \textbf{0.00} & 0.78 & 0.04 & 0.25 & 2.37\\
     & 22,000 & 0.71 & 1.92 & 3.74 & \textbf{0.00} & 0.28 & 1.02 & 0.07 & 0.67 & 2.65\\
     & 94,000 & \textbf{0.50} & \textbf{1.32} & \textbf{2.61} & \textbf{0.00} & 0.04 & \textbf{0.71} & \textbf{0.00} & \textbf{0.21} & \textbf{1.94}\\
    \bottomrule
  \end{tabular}
  \vspace{3pt}
  \caption{Evaluation of planning metrics on nuScenes and ONCE by adding the occupancy cost maps to residual cost maps during training.}
  \label{tab:once}
\end{table*}

\newtext{
\subsection{Limitations}

We highlight a few limitations of our work.
First, our self-supervised emergent occupancy does not offer semantics (e.g., traffic light and lane information) that is crucial for urban navigation. Despite this, we show that learning to drive with future occupancy is a safe fallback option in industrial autonomous driving.
Second, BEV occupancy by itself does not handle overhead structures (e.g., trees, overpass); this may be mitigated by learning which occupied voxels are `passable' during differentiable raycasting.
Third, we rely on open-loop evaluation, where the world (incorrectly) unfolds in the same manner as the expert trajectory. Although this can be corrected in a closed-loop setup, with our work we show that optimizing for collision metrics, with or without L2 error, can act as a proxy for learning to drive safe in real-world.
Fourth, our method assumes accurate ego-motion during training but does not require it at test time.
Finally, as the supplementary video highlights, our occupancy estimates diffuse over time, capturing multiple futures. However, we posit that future occupancy can be made more robust by constraining it with scene flow.
}

{\small
\bibliographystyle{splncs04}
\bibliography{egbib}
}

\end{document}